\documentclass[a4paper,conference]{IEEEtran}
\ifCLASSINFOpdf
\else
\fi

\usepackage{graphicx}
\usepackage{comment}
\usepackage{amsmath,amssymb} 
\usepackage{color}
\usepackage{url}

\usepackage{times}
\usepackage{epsfig}

\usepackage{booktabs}
\usepackage{tabularx}
\usepackage{multirow}
\usepackage{subfigure}
\usepackage{dsfont}
\usepackage{array}
\usepackage{gensymb}
\usepackage{rotating}
\usepackage{pifont}
\usepackage{cite}

\usepackage{verbatim}
\usepackage{xcolor}

\newcommand{\cmark}{\ding{51}}
\newcommand{\specialcell}[2][l]{\begin{tabular}[#1]{@{}l@{}}#2\end{tabular}}

\newcommand{\ie}[0]{\textit{i.e.}}
\newcommand{\eg}[0]{\textit{e.g.}}
\newcommand{\etal}[0]{\textit{et al.}}

\begin{document}
%
\title{Learn to Segment Retinal Lesions and Beyond}

\author{\IEEEauthorblockN{
Qijie Wei\IEEEauthorrefmark{1}\IEEEauthorrefmark{2}, 
Xirong Li\IEEEauthorrefmark{1}\IEEEauthorrefmark{2}, 
Weihong Yu\IEEEauthorrefmark{3}, 
Xiao Zhang\IEEEauthorrefmark{3}, 
Yongpeng Zhang\IEEEauthorrefmark{4}, 
Bojie Hu\IEEEauthorrefmark{5}, 
Bin Mo\IEEEauthorrefmark{4}\\ 
Di Gong\IEEEauthorrefmark{6}, 
Ning Chen\IEEEauthorrefmark{7}, 
Dayong Ding\IEEEauthorrefmark{2}, 
Youxin Chen\IEEEauthorrefmark{3}}

\IEEEauthorblockA{\IEEEauthorrefmark{1}Key Lab of DEKE, Renmin University of China, Beijing, China}
\IEEEauthorblockA{\IEEEauthorrefmark{2}Vistel AI Lab, Visionary Intelligence Ltd, Beijing, China}
\IEEEauthorblockA{\IEEEauthorrefmark{3}Peking Union Medical College Hospital, Beijing, China}
\IEEEauthorblockA{\IEEEauthorrefmark{4}Beijing Tongren Hospital, Beijing, China}
\IEEEauthorblockA{\IEEEauthorrefmark{5}Tianjin Medicial University Eye Hospital, Tianjin, China}
\IEEEauthorblockA{\IEEEauthorrefmark{6}China-Japanese Riendship Hospital, Beijing, China}
\IEEEauthorblockA{\IEEEauthorrefmark{7}The Affilliated Yantai Yuhuangding Hospital of Qingdao University, Yantai, China\\
Email: qijie.wei@vistel.cn, xirong@ruc.edu.cn, chenyouxinpumch@163.com}
}


%


\maketitle

\begin{abstract}
Towards automated retinal screening, this paper makes an endeavor to simultaneously achieve pixel-level retinal lesion segmentation and image-level disease classification. Such a multi-task approach is crucial for accurate and clinically interpretable disease diagnosis. Prior art is insufficient due to three challenges, \ie, lesions lacking objective boundaries, clinical importance of lesions irrelevant to their size, and the lack of one-to-one correspondence between lesion and disease classes. This paper attacks the three challenges in the context of diabetic retinopathy (DR) grading. We propose \emph{Lesion-Net}, a new variant of fully convolutional networks, with its expansive path re-designed to tackle the first challenge. A dual Dice loss that leverages both semantic segmentation and image classification losses is introduced to resolve the second challenge. Lastly, we build a multi-task network that employs Lesion-Net as a side-attention branch for both DR grading and result interpretation. A set of 12K fundus images is manually segmented by 45 ophthalmologists for 8 DR-related lesions, resulting in 290K manual segments in total. Extensive experiments on this large-scale dataset show that our proposed approach surpasses the prior art for multiple tasks including lesion segmentation, lesion classification and DR grading.
\end{abstract}


%
\IEEEpeerreviewmaketitle

\section{Introduction} \label{sec:intro}

Given the increasing demand of retinal screening and the clear shortage of experienced ophthalmologists, fundus image based retinal disease diagnosis is crucial for the well-being of many~\cite{referable,miccai19-multidis,miccai19-amd}.
Previous studies on fundus image segmentation concentrate on anatomical structures in retina including optic disc / cup and vessels \cite{tmi18-mnet,tmi19-cenet,cvprw19-m2u-net}. By contrast, this paper aims for \emph{retinal lesions}, which are symptoms of ocular fundus diseases manifested in color fundus images. By answering the question of what lesions are in a fundus image and where in the image they are located, lesion segmentation has a potential to enable clinical interpretability of disease classes predicted at the image level. Attacking lesion segmentation and retinal disease classification in a unified framework is thus valuable.

Note that for natural images as in PASCAL-VOC alike tasks \cite{wildcat,Ge_2018_CVPR,Zhou_2018_CVPR}, the semantic segmentation task and the image classification task typically share the same class vocabulary. Consequently, developing a multi-task approach seems to be relatively straightforward, \eg, by converting classes predicted at the pixel-level to the image-level by max or mean pooling.
For fundus images, however, lesion labels and disease classes are distinct and lack one-to-one correspondence. See for instance lesions used in the clinical practice guidelines for diabetic retinopathy\footnote{Diabetic retinopathy is a complication of diabetes mellitus caused by damage to blood vessels of the light-sensitive tissue at the retina \cite{dc17-dr}.} (DR) grading in Table \ref{tab:DRLevel}.
This means lesion segmentation cannot be directly converted to image-level DR grades. 
Hence, a unified framework that  effectively segments lesions and exploits the segmentation for accurate disease classification is in demand.


\begin{table} [tb!]
\renewcommand{\arraystretch}{1}
\centering
\caption{\textbf{A summary of the American Academy of Ophthalmology (AAO) preferred practice pattern guidelines for diabetic retinopathy grading}. As venous beading and IrMA are very difficult to be recognized even for ophthalmologists and occur rarely, we exclude them from this study. The eight lesions stuided in this work are indicated by \cmark.}
\label{tab:DRLevel}
\scalebox{0.9}{
\begin{tabular}{@{} l |l | l @{}}
\toprule
\textbf{Grade} & \multicolumn{2}{c}{\textbf{Lesion evidence for DR grading}} \\
\hline
               & \emph{Sufficient} & \emph{Indirect} \\
\hline
DR1 & $\bullet$ Microaneurysm (MA), exclusively \cmark & -- \\
\hline
DR2 & $\bullet$ Intraretinal hemorrhage (iHE) \cmark & $\bullet$ Hard exudate (HaEx) \cmark \\
\hline
DR3 & \specialcell{Any of the following:\\ 
                     $\bullet$ Over 20 iHEs in each of 4 quadrants \\
                     $\bullet$ Venous beading in 2+ quadrants \\
                     $\bullet$ IrMA in 1+ quadrants} & \specialcell{$\bullet$ Cotton-wool spot \\ (CWS) \cmark} \\ 
\hline
DR4 & \specialcell{Any of the following:\\
                     $\bullet$ Neovascularization (NV) \cmark \\
                     $\bullet$ Vitreous hemorrhage (vHE) \cmark \\
                     $\bullet$ Preretinal hemorrhage (pHE) \cmark} & \specialcell{$\bullet$ Fibrous proliferation\\ (FiP) \cmark} \\

\bottomrule
\end{tabular}}
\vspace{-6mm}
\end{table}



\begin{figure} [tb!]
\centering
\includegraphics[width=\columnwidth]{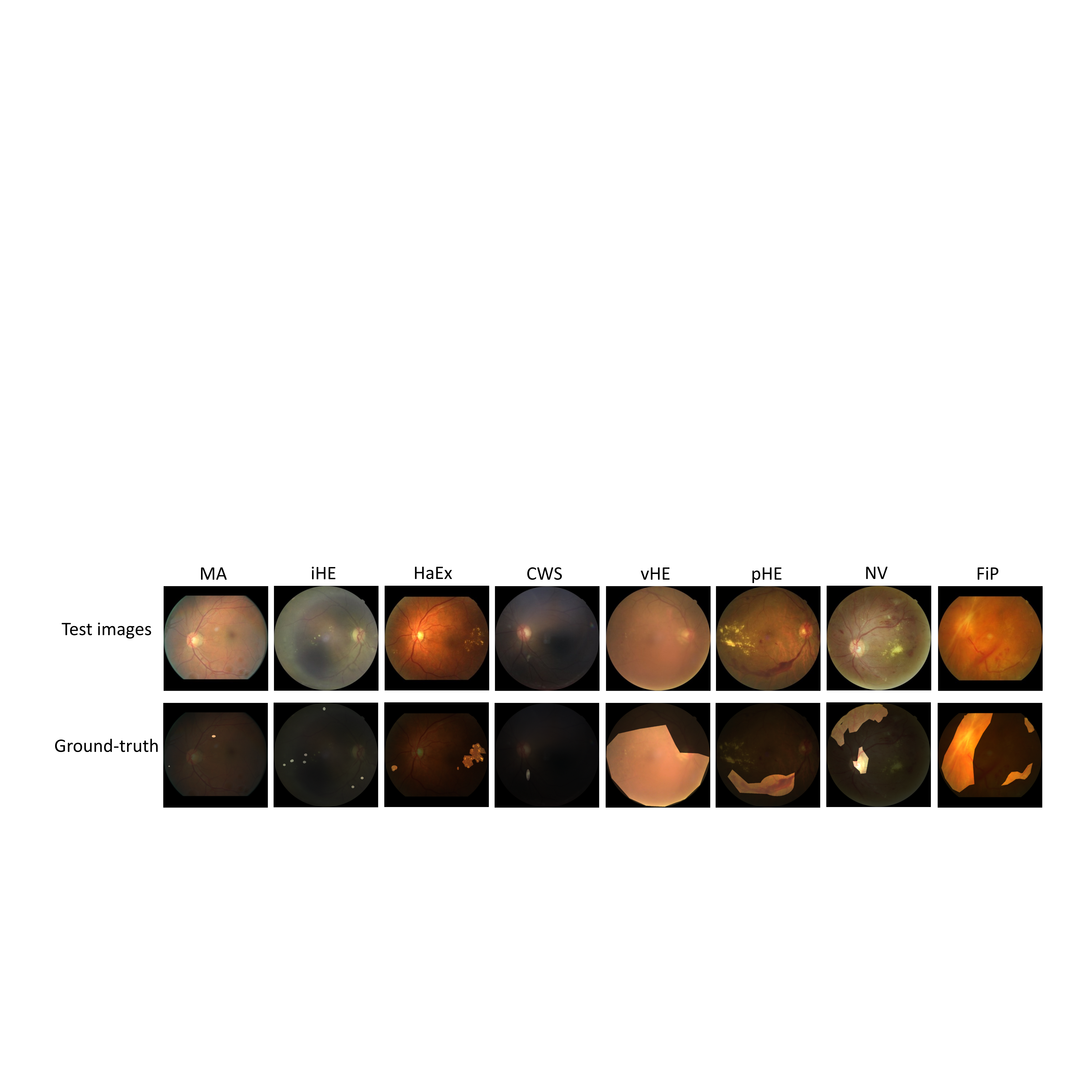}
\caption{
\textbf{Visual examples of 8 DR-related retinal lesions studied in this paper}. For a clear view, we show only one lesion per image.}
\label{fig:example}
\vspace{-6mm}
\end{figure}

Given a fundus image, instances of a specific lesion class occupy a specific region or multiple regions with diverse visual appearance, see Fig. \ref{fig:example}. 
With the advent of fully convolutional networks (FCN) \cite{fcn}, exciting progress has been made in semantic segmentation, especially for natural scenes~\cite{dialated,segnet,deeplabv3p,cvpr19-danet,iccv19-ccnet}. However, directly applying the state-of-the-art for retinal lesion segmentation is problematic. 
Unlike objects in natural images, retinal lesions lack clear boundaries against the background. It is practically impossible for ophthalmologists to segment lesions at the same preciseness, meaning an FCN has to learn from annotations with imprecise boundaries. In the meanwhile, for diagnosis, it is mostly the presence and locality of specific lesions that are involved, see Table \ref{tab:DRLevel}. 
Extremely precise segmentation is not only difficult to achieve but also unnecessary from a clinical view. We thus hypothesize that cutting-edge FCNs, \eg, DeepLabv3+~\cite{deeplabv3p}, are over-designed for lesion segmentation.

Moreover, while the importance of an object in a natural image is largely reflected by its size~\cite{cvpr12-berg-import}, the importance of a lesion in a fundus image does not count on the amount of pixels it possesses. Diabetic retinopathy, depending on what lesions are presented, is categorized into five levels, from DR0 (\ie, no DR) to DR4~(\ie, proliferative DR). 
The presence of a preretinal hemorrhage, even though in a relatively small size, means DR4. Such a property cannot be well addressed by current segmentation losses including cross entropy \cite{fcn,unet,cvpr18-seg-every}, focal loss~\cite{iccv17-focal,Yu_2018_CVPR}, and Dice \cite{vnet,tmi18-mnet}.


To conquer the aforementioned challenges, this paper makes the first endeavor to simultaneously solve retinal lesion segmentation and disease classification in an end-to-end framework. We choose DR, a leading cause of blindness~\cite{dc17-dr}, as our target disease. Our main novelties are


$\bullet$ We study eight lesions including microaneurysm (MA), intraretinal hemorrhage (iHE), hard exudate (HaEx), cotton-wool spot (CWS), vitreous hemorrhage (vHE), preretinal hemorrhage (pHE), neovascularization (NV), and fibrous proliferation (FiP) that support the full range of DR grades. This is a new state-of-the-art in terms of quantity, complexity and clinical usability.

$\bullet$ We propose Lesion-Net for retinal lesion segmentation. While inheriting FCN's classical contracting-and-expansive structure, Lesion-Net has a re-designed expansive path with its length adjustable and its upsampling operation lightweight trainable. We adopt a \emph{dual loss} that combines both semantic segmentation and image classification losses. These two designs enable Lesion-Net to effectively learn from lesion annotations with imprecise boundaries and to substantially reduce false alarms of small-size lesions. 



$\bullet$ We propose a multi-task network that effectively harnesses lesion segmentation maps, as side information, for improving DR grading. Such an attention mechanism conceptually differs from prevalent self-attention mechanisms~\cite{nips17-atten,abn,cvpr19-danet}. Once trained, the multi-task network performs three tasks, \ie, lesion segmentation, lesion classification and DR grading, all in one forward pass. 


$\bullet$ We conduct extensive experiments on 12K color fundus images collected from Kaggle \cite{kaggle} and local hospitals. With 290K expert-labeled pixel-level lesion segments, the dataset is the largest of its kind. The experiments confirm the superiority of both Lesion-Net and the multi-task network against the prior art including FCN~\cite{fcn}, U-Net~\cite{unet}, DANet~\cite{cvpr19-danet}, and DeepLabv3+~\cite{deeplabv3p} for lesion segmentation and classification, Inception-v3~\cite{inception-v3} and ABN~\cite{abn} for DR grading. 
To promote related research, we have released the Kaggle part of our test data, containing 1,593 images and 34,268 expert-labeled lesion segments\footnote{\url{https://github.com/WeiQijie/retinal-lesions}}, substantially larger than the present-day dataset that has only 81 images with manual segmentation for four lesions~\cite{mia20-idrid}.




\section{Related Work} \label{sec:related}

\textbf{Models for semantic segmentation}. 
Since Long \etal \cite{fcn}, FCNs have been the \emph{de facto} standard technique for semantic segmentation. An FCN can be conceptually decomposed into a contracting path and an expansive path. The contracting path progressively extracts and downsamples feature maps from an input image. The expansive path, by transforming and upsampling, produces a full-resolution segmentation map of the same size as the input image. Towards more precise segmentation, novel designs are continuously proposed either in the contracting path, or in the expansive path or in both. For instance, dilated convolutions are introduced in \cite{dialated}, so the contracting path can produce feature maps with higher resolutions to preserve more detailed spatial information. In U-Net \cite{unet}, the contracting path and the expansive path are carefully designed to be symmetrical. Skip connections from the contracting path to the expansive path are added, again for the purpose of preserving spatial information to generate more accurate segmentation boundaries. In order to capture long-range contextual information in both spatial and channel dimensions, DANet \cite{cvpr19-danet} introduces a position attention module and  a channel attention module in the expansive path. The state-of-the-art DeepLabv3+ uses both dilated convolutions and spatial pyramid pooling in its contracting path \cite{deeplabv3p}.  Its expansive path uses multiple skip connections to exploit features from lower levels. As identifying the precise boundary of a retinal lesion is secondary to the practical use of lesion segmentation, a new FCN is required.


\textbf{Retinal lesion segmentation}. While earlier works for retinal lesion segmentation use traditional image processing techniques~\cite{fleming2010role,niemeijer2007automated}, current works mostly take a patch-based deep learning approach \cite{tmi16-hemorrhage-detection,ma-he-exudates,high-risk-lesions,miccai18-lesion-detction}. In \cite{tmi16-hemorrhage-detection}, for instance, a customized CNN is used to segment iHE by patch classification. Similarly in \cite{high-risk-lesions}, a patch-trained CNN is applied in a sliding window manner, classifying every grid in a test image into five classes, \ie, normal, MA, iHE, HaEx and high-risk lesion.  By predicting whether a given patch contains a specific lesion, segmentation maps obtained by the above works tend to be sparse and imprecise. A more fundamental drawback is that the approach lacks a holistic view. Consider MA and iHE for instance. The two lesions are visually close as both are small lesions look like dark dots. However, MA occurs around vessels. Also, an image with no other lesion is more likely to have MA than iHE. For a model looking only at local areas, modeling these kinds of holistic clues is difficult. 

\textbf{Lesion-enhanced DR grading}. 
Initial efforts have been made towards lesion-enhanced DR grading. A two-step method is developed in \cite{lesion-dr1}, where an input image is first converted to a weight map by using a CNN to classify all patches of the image as normal, MA or iHE. The image, multiplied by the weight map, is fed into a DR grading network. A lesion-guided attention mechanism is described in \cite{lesion-dr2} to weigh specific regions in the input image. Three lesions are considered: MA, iHE and HaEx. Neither of these works considers severe lesions such as pHE, vHE, and NV. 


\textbf{Attention-enhanced image classification}. The state-of-the-art is Attention Branch Network (ABN) \cite{abn}, which extends a response-based visual explanation model~\cite{cvpr16-cam} by introducing an attention branch into a specific CNN. Consequently, ABN not only improves image classification but also produces an attention map to interpret the decision. Note that the attention is self-generated. Our attention mechanism exploits the output of the semantic segmentation network as side information, and thus conceptually differs from ABN.

\section{Approach} \label{sec:approach}

Given a color fundus image, we aim to perform lesion segmentation, classification and subsequently DR grading in a unified framework. We use $\mathcal{X}$ to denote a specific $s \times s$ image, which contains an array of $s^2$ pixels $\{x_1,\ldots,x_i,\ldots,x_{s^2}\}$. Let $\mathcal{L}=\{l_1,\ldots,\l_m\}$ be $m$ lesions in consideration. Regions of distinct lesions, \eg, HaEx and iHE, often overlap partially, meaning a pixel can be assigned with multiple labels. So the goal of lesion segmentation is to automatically assign to each pixel $x_i$  a $m$-dimensional probabilistic vector, $\mathbf{p}_i=\{p_{i,1}, \ldots, p_{i,m}\}$, where $p_{i,j} \in [0,1]$ indicates the probability of the pixel belonging to the $j$-th lesion. 
%
Lesion classification is to predict lesions at the image level. Given the probabilistic segmentation map $\{\mathbf{p}_1, \ldots, \mathbf{p}_{s^2}\}$, the probability of the presence of a specific lesion $l_j$, denoted as $P_j$, is naturally obtained by global max pooling on the map, \ie,
\begin{equation} \label{eq:Pj}
P_j := \max\{p_{1,j}, \ldots, p_{s^2,j} \},~~~j=1,\ldots,m.
\end{equation}
For both lesion segmentation and classification, hard labels are obtained by thresholding at $0.5$.
As for DR grading, the goal is to exclusively assign one of the following labels, \ie, \{DR0, DR1, DR2, DR3, DR4\}, to the given image. 

Next, we  depict the proposed lesion segmentation network, followed by the multi-task network. 

\subsection{Lesion-Net for Retinal Lesion Segmentation} \label{ssec:lesion-seg-net}


\begin{figure*} [tb!]
\centering
\includegraphics[width=1.9\columnwidth]{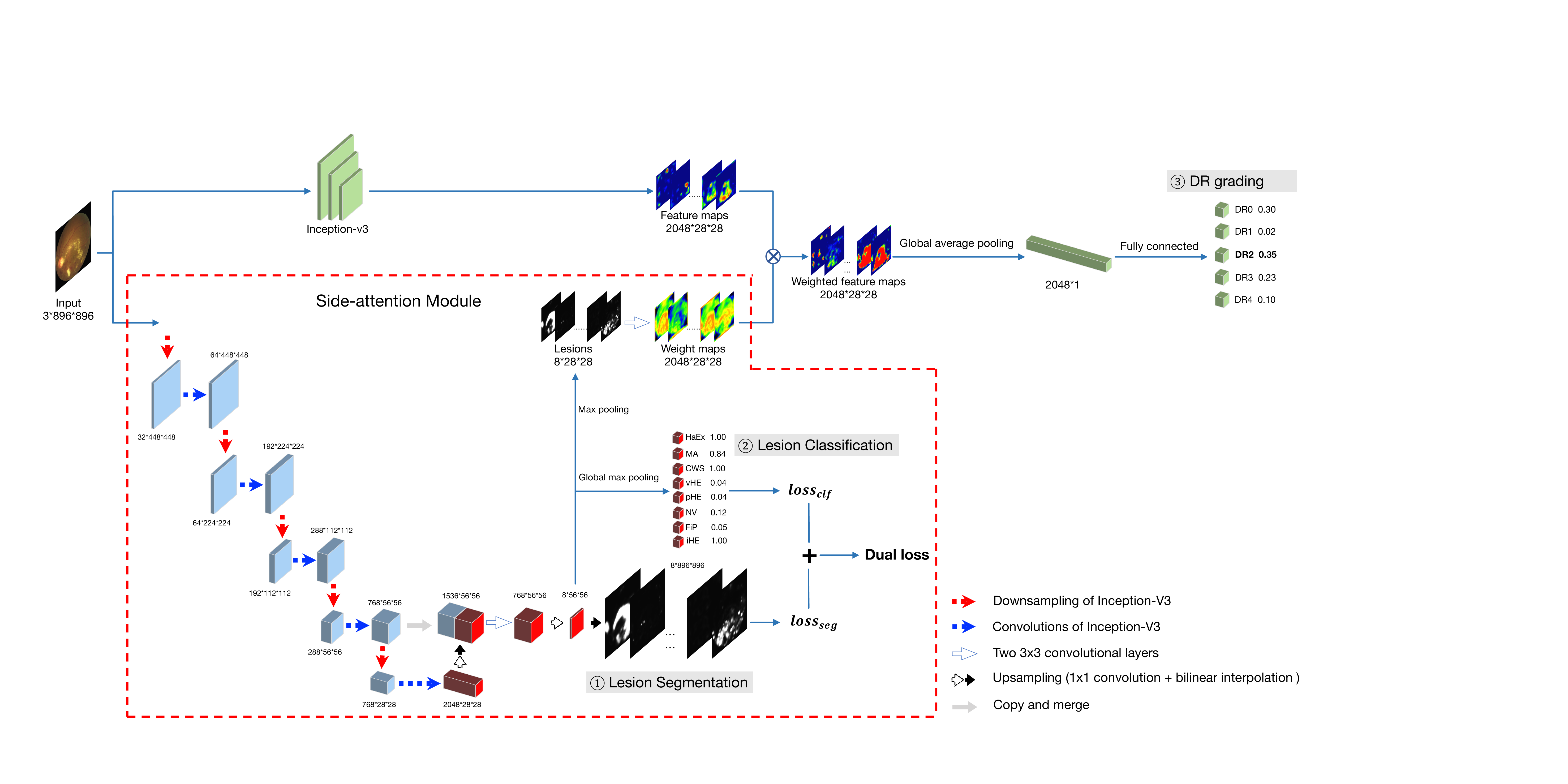}
\caption{
\textbf{Proposed multi-task network for (1) lesion segmentation, (2) lesion classification and (3) DR grading}. Given a color fundus image, 
Lesion-Net-16s (the lower branch) generates probabilistic segmentation maps for eight lesions. Lesion classification is accomplished by global max pooling on the maps. For lesion-enhanced DR grading, a side-attention branch is used to fuse the segmentation maps with an array of 2,048 feature maps from  Inception-v3 in the upper branch. Compared with directly weighing the feature maps with the segmentation maps, the trainable side-attention is more effective.
}
\label{fig:l-net}
\vspace{-6mm}
\end{figure*}

\begin{figure} [tb!]
\centering
\includegraphics[width=1\columnwidth]{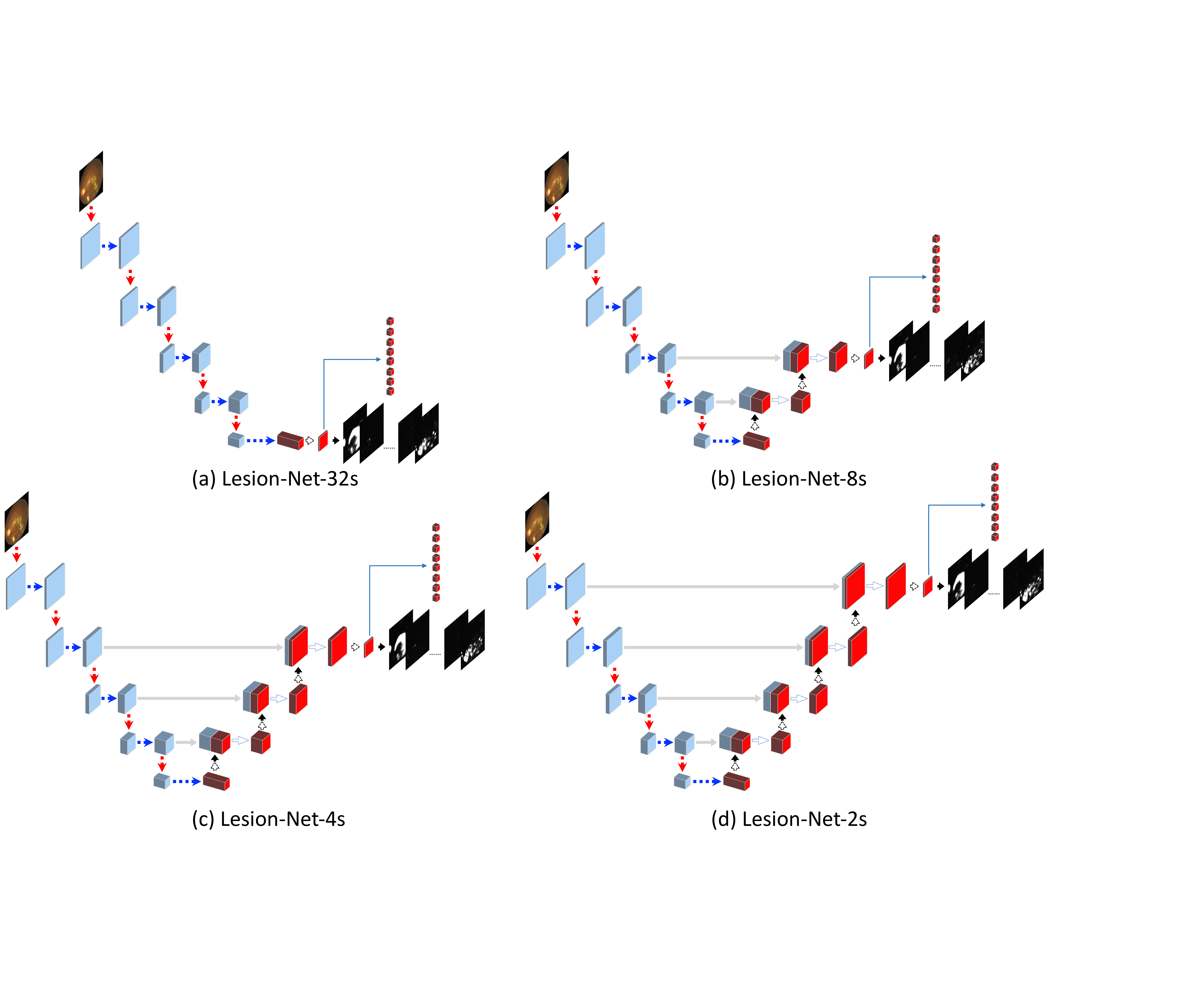}
\caption{
\textbf{Variants of Lesion-Net}. The variable-length expansive path enables learning from lesion annotations with imprecise boundaries.}
\label{fig:l-net-settings}
\vspace{-7mm}
\end{figure}

\textbf{Network architecture}. 
For the contracting path of Lesion-Net, we use convolutional blocks of Inception-v3 \cite{inception-v3} for its outstanding feature extraction ability. Note that other state-of-the-art CNNs \cite{resnet,densenet,icml19-efficientnet} can, in principle, be used.

Our task-specific design lies in the expansive path, where we leverage the effectiveness of U-Net \cite{unet} for re-using information from the contracting path and the flexibility of the original FCN \cite{fcn} in cutting off the expansive path for preventing over-precise segmentation. 

In concrete, in order to  re-use feature maps from the contracting path, we adopt U-Net's copy-and-merge strategy instead of adding operations in the FCN, see Fig. \ref{fig:l-net}. For upsampling, we replace U-Net's deconvolution by a $1 \times 1$ convolution to adjust the number of feature maps and subsequently a parameter-free bilinear interpolation to enlarge the feature maps. Such a tactic not only reduces the number of parameters. By applying an element-wise sigmoid activation, the output of the $1 \times 1$ convolution is naturally transformed to $m$ probabilistic maps with respect to the $m$ lesions.

The fact that retinal lesions lack accurate boundaries makes it unnecessary to seek for very precise segmentation. While the symmetry between the contracting and expansive paths in U-Net is useful in its original context of cell segmentation, we argue that such a constraint is unnecessary for the current task. In fact, extra parameters introduced by the symmetry into the expansive path increases the difficulty of training the network. Therefore, we let the length of Lesion-Net's expansive path adjustable. If the expansive path is cut at an early stage with feature maps of size $28 \times 28$, the maps need to be upsampled by a factor of $32$ to produce the final segmentation maps. Following the convention of \cite{fcn}, we term this variant Lesion-Net-32s. By contrast, Lesion-Net-2s exploits all the intermediate feature maps. The models that fall in between are Lesion-Net-16s, Lesion-Net-8s and Lesion-Net-4s. 
Fig. \ref{fig:l-net-settings} shows Lesion-Net with distinct expansive paths.

\textbf{Loss function}. 
Training Lesion-Net is nontrivial due to the following two issues.
First, while the area of a specific lesion varies, the importance of the lesion does not depend on its size. This property cannot be well reflected in a pixel-wise loss, to which a smaller blob contributes less. 
Misclassifying a small blob does not lead to a significant increase in the segmentation loss, and thus difficult to be corrected during training. Such a small misclassification, even though ignorable from the viewpoint of semantic segmentation, can be crucial for proper diagnosis of related diseases.
Second, the data is extremely imbalanced, making commonly used loss functions such as cross entropy ineffective. Our study on a set of 12k expert-labeled fundus images shows that pixels of lesions account for less than 1\%. By contrast, for PASCAL VOC2012 \cite{ijcv15-voc}, a popular benchmark set for natural image segmentation, the proportion of pixels corresponding to objects is about 25\%. We find in preliminary experiments that with the cross-entropy loss, the lesion segmentation model easily got trapped in a local optimum, predicting all pixels as negative, albeit a very low training loss.

To jointly address the two issues, we introduce a new \emph{dual loss} that combines a semantic segmentation loss $loss_{seg}$ and an image classification loss $loss_{clf}$, \ie,
\begin{equation}\label{eq:dual-loss}
loss_{dual} = \lambda \cdot loss_{seg} + (1 - \lambda) \cdot loss_{clf},
\end{equation}
where $\lambda \in [0,1]$ is a hyper parameter to strike a balance between the two sub-losses. 
We instantiate both $loss_{seg}$ and $loss_{clf}$
using the Dice loss, previously used for segmenting prostate MRI~\cite{vnet} and optic disc / cup~\cite{tmi18-mnet}. Our ablation study in Section \ref{sssec:exp-lesion-seg} shows that Dice loss is more effective than Weighed Cross Entropy~\cite{gdice} and Focal Loss~\cite{iccv17-focal}. The weight $\lambda$ is empirically set to $0.8$ based on a held-out validation set, a common practice for selecting hyper parameters. 

Note that the dual Dice loss is conceptually different from the multi-scale Dice \cite{tmi18-mnet}, which combines pixel-level Dice losses computed from images of varied scales and thus remains insensitive to small-sized errors. The motivation of 
our dual loss also differs from the combined cross-entropy loss described in \cite{zhang2018unified}, where the image classification loss is used as a regularization term to reduce overfitting. 

Given a mini-batch of $n$ images, we compute the Dice version of $loss_{seg}$ as
\begin{equation} \label{eq:dice-pix}
loss_{seg} = 1 - \dfrac{2 \cdot \sum_{i=1}^{n \cdot s^2}{\sum_{j=1}^{m}{p_{i,j} \cdot t_{i,j}}}}{\sum_{i=1}^{n \cdot s^2}{\sum_{j=1}^{m}{p_{i,j}^2}} + \sum_{i=1}^{n \cdot s^2}{\sum_{j=1}^{m}{t_{i,j}^2}}},
\end{equation}
where $t_{i,j} \in \{0,1\}$ is ground truth of the $i$-th pixel with respect to the $j$-th lesion. In the extreme case where all pixels are predicted as negative, the dice loss is close to $1$.

We compute the Dice version of $loss_{clf}$ as 
\begin{equation} \label{eq:dice-img}
loss_{clf} = 1 - \dfrac{2 \cdot \sum_{i=1}^{n}{\sum_{j=1}^{m}{P_{i,j} \cdot T_{i,j}}}}{\sum_{i=1}^{n}{\sum_{j=1}^{m}{P_{i,j}^2}} + \sum_{i=1}^{n}{\sum_{j=1}^{m}{T_{i,j}^2}}},
\end{equation}
where $T_{i,j} \in \{0,1\}$ is the ground-truth label indicating whether the $j$-th lesion is present in the $i$-th image in the given batch. Recall that both $P_{i,j}$ and $T_{i,j}$ are obtained by global max pooling on the pixel-level labels, so $P_{i,j}$, $T_{i,j}$ and accordingly $loss_{clf}$ are all invariant to the lesion size.


%
%
%


\subsection{Multi-task Network for Lesion-enhanced DR Grading} \label{ssec:grading-net}

To predict DR grades, we choose Inception-v3 \cite{inception-v3} as our baseline model. This model has established the state-of-the-art for predicting referable DR \cite{referable}, age-related eye diseases \cite{amd-deep} and other retinal abnormalities~\cite{miccai19-multidis}. In fact, for all models evaluated in this work, we use Inception-v3 as their backbones for fair comparison. To produce a probabilistic score per DR grade, we modify Inception-v3 by adding after the global average pooling (GAP) layer a fully connected layer of size $2,048 \times 5$, followed by a softmax layer.  Different from previous works that use a typical resolution of $299\times 299$ \cite{referable}, we use a much larger resolution of $896 \times 896$, making our Inception-v3 a much stronger baseline.






For lesion-enhanced DR grading, we propose a multi-task network, with its overall architecture shown in Fig. \ref{fig:l-net}. The multi-task network consists of two branches. At the top is its main branch, with Inception-v3 as the backbone, that performs DR grading. The side-attention branch, with Lesion-Net as its backbone, is responsible for injecting semantic and spatial information contained in the $m$ lesion segmentation maps into the main branch. In particular, the injection is performed at the last feature maps, denoted as $\{f_1,\ldots,f_{k}\}$, in the main branch, with $k=2,048$. To that end, the side-attention branch shall generate the same number of weight maps, denoted as $\{w_1,\ldots,w_{k}\}$. Multiplying the feature maps by the weight maps side by side generates new weighted feature maps as 
\begin{equation}
\{f_1 \otimes w_1, f_2 \otimes w_2, \ldots, f_{k} \otimes w_{k}\},
\end{equation}
where $\otimes$ indicates element-wise multiplication. The new feature maps then go through a GAP layer, followed by a classification block. 
It is worth point out that the weight information is essentially from the side-attention branch rather than generated by the main branch itself. Hence, the multi-task network is conceptually different from self-attention networks~\cite{nips17-atten,abn,cvpr19-danet}.




To convert the lesion segmentation maps $\{s_1, \ldots, s_m\}$ into the weight maps, an intuitive strategy is to let the main branch pay attention to regions with maximal lesion response. This is achieved by channel-wise max pooling (CW-MaxPool) over the segmentation maps\footnote{The segmentation maps here are already down-sampled to the same size as the feature maps in the main branch.}, \ie
\begin{equation} \label{eq:att-w1}
w_i := \mbox{CW-MaxPool}(\{s_1, \ldots, s_m\}),~~~i=1,\ldots,k.
\end{equation} 
However, a region deemed to be negative with respect to the lesions does not necessarily mean it is useless for DR grading. So we further consider a learning based strategy, using a lightweight convolutional block consisting of two $3\times 3$ convolutional layers, \ie
\begin{equation} \label{eq:att-w2}
\{w_1,\ldots,w_k\} := \mbox{Conv}(\{s_1, \ldots, s_m\}).
\end{equation}
We train the multi-task network with the cross-entropy loss, commonly used for multi-class image classification.





\section{Evaluation} \label{sec:experiment}

\subsection{Experimental Setup} \label{ssec:setup}


According to the AAO guidelines \cite{aao2017}, there are seven lesions used as sufficient evidence for specific DR grades. Among them, venous beading and IrMA are very difficult to be recognized even for ophthalmologists and occur rarely. So they are excluded from this study. We include three other lesions, \ie,~HaEx, CWS, and FiP, indirectly related to DR grading. 
We compile a final list of eight lesions, see Table \ref{tab:DRLevel}.

\textbf{Ground-truth construction}. Public datasets suited for our purpose does not exist. So we construct a large collection of 12,252 color fundus images with both pixel-level lesion annotations and image-level DR grades as follows. 
We collected initially 23K color fundus images of posterior pole, consisting of 12k images from our hospital partners and 11k images randomly sampled from the Kaggle DR Detection task \cite{kaggle}. While the images were from patients with diabetes, some of them show other eye diseases such as glaucoma, AMD and RVO. So DR0 does not necessarily mean a healthy eye. Such a characteristic makes the data close to the real scenario and thus challenging. 

For expert labeling, a panel of 45 experienced ophthalmologists was formed. We developed a web-based annotation system, where an annotator marks out  lesions in a given image using either ellipses or polygons and accordingly grade the image. Lesion annotation and DR grading from a single image are somewhat subjective. So for quality control, each image was assigned to at least three annotators. Images receiving consistent DR grades, \ie,~the majority vote for a specific grade, are preserved. Accordingly, per image we cleaned lesion annotations so they are complied to the diagnostic guidelines. Eventually, we obtain 12,252 images with 290K expert-labeled lesion segments. We split the dataset at random for training (70\%), validation (10\%) and test (20\%). 




\textbf{Implementations}. An input image is sized to $896 \times 896$, as small lesions can not be seen well in lower resolution. We use SGD with a weight decay factor of $0.0001$ and a momentum of $0.95$. The initial learning rate is $0.001$. Validation occurs every 1K batches. If the validation performance does not improve in 4 consecutive validations, the learning rate will be divided by $10$. Early stop occurs once the performance does not improve in 10 consecutive validations. 
For training Lesion-Net, 
we start with $loss_{seg}$. Once the learning rate is reduced, $loss_{seg}$ is replaced by $loss_{dual}$. For DR grading, a pre-trained Lesion-Net is used for the multi-task network. We tried to train both branches, but found no improvement in DR grading yet an absolute decrease of 0.01 in the segmentation performance. So we did not go further in that direction. For the varied models assessed in this paper, we use Inception-v3 pre-trained on ImageNet~\cite{cvpr09-imagenet} as their backbones. Random rotation, crop, flip and random changes in brightness, saturation and contrast are used for data augmentation. Training was performed using PyTorch on two NVIDA Tesla P40 GPUs. 

\textbf{Evaluation criteria}. For lesion segmentation, we report pixel-wise F1 score, the harmonic mean of precision and recall. For lesion classification, we report image-wise F1. As lesions predicted at the pixel level are propagated to the image level via global max pooling, the two criteria complements each other, providing a more comprehensive assessment of a specific segmentation model. For DR grading, We report the quadratic weighted \emph{kappa}, which measures inter-annotator agreement and used by the Kaggle DR Detection task~\cite{kaggle}. 



\subsection{Experiment 1. Lesion Segmentation} \label{sssec:exp-lesion-seg}

\begin{table*} [tb!]
\renewcommand{\arraystretch}{1.1}
\centering
\caption{\textbf{Lesion segmentation and classification by different models}.}
\label{tab:lesion-seg}
\scalebox{0.75}{
\begin{tabular}{@{}ll r r r r r r r r r r r r r r r r r r r@{}}
\toprule
\multirow{1}{*}{\textbf{Model}} && \multicolumn{9}{c}{Lesion segmentation} && \multicolumn{9}{c}{Lesion classification} \\
\cmidrule{3-11} \cmidrule{13-21}
&& \textbf{Mean} & MA    & iHE   & HaEx  & CWS   & vHE   & pHE   & NV & FiP && \textbf{Mean} & MA    & iHE   & HaEx  & CWS   & vHE   & pHE   & NV & FiP\\
\cmidrule{1-1} \cmidrule{3-11} \cmidrule{13-21}
patch FCN-32s       && 0.553          & 0.209          & 0.583          & 0.714          & 0.535          & 0.622          & 0.549          & 0.554          & 0.659 
                    && 0.704          & 0.886          & 0.849          & 0.828          & 0.720          & 0.634          & 0.544          & 0.637          & 0.535\\
FCN-32s             && 0.571          & 0.327          & 0.592          & 0.728          & 0.528          & 0.642          & 0.562          & 0.530          & 0.662  
                    && 0.769          & 0.900          & 0.858          & 0.856          & 0.771          & 0.722          & 0.683          & 0.694          & 0.669\\
FCN-16s             && 0.587          & 0.369          & 0.608          & 0.737          & 0.575          & 0.639          & 0.515          & 0.581          & 0.671 
                    && 0.787          & 0.890          & 0.849          & 0.847          & 0.743          & 0.758          & 0.726          & 0.696          & 0.783\\
FCN-8s              && 0.586          & 0.369          & 0.609          & 0.740          & 0.573          & 0.640          & 0.534          & \textbf{0.583} & 0.639  
                    && 0.778          & 0.891          & 0.858          & 0.854          & 0.749          & 0.766          & 0.711          & 0.671          & 0.725\\

U-Net               && 0.570          & 0.384          & 0.598          & 0.730          & 0.565          & 0.547          & 0.604          & 0.538          & 0.592  
                    && 0.757          & 0.888          & 0.855          & 0.843          & 0.755          & 0.639          & 0.689          & 0.653          & 0.737\\

DeepLabv3+          && 0.553          & 0.367          & 0.612          & 0.732          & 0.558          & 0.550          & 0.477          & 0.498          & 0.631  
                    && 0.794          & 0.899          & 0.863          & 0.866          & 0.764          & \textbf{0.800} & 0.693          & 0.677          & 0.792\\ 
DANet               && 0.585          & 0.351          & 0.608          & 0.733          & 0.560          & 0.623          & 0.589          & 0.543          & 0.671 
                    && 0.775          & 0.900          & 0.853          & 0.852          & 0.772          & 0.713          & 0.682          & 0.715          & 0.712\\ [3pt]

Inception-v3        && --             & --             & --             & --             & --             & --             & --             & --             & --
                    && 0.716          & 0.895          & 0.893          & 0.865          & 0.766          & 0.500          & 0.540          & 0.594          & 0.678 \\ 
ABN-lesion          && --             & --             & --             & --             & --             & --             & --             & --             & --
                    && 0.726          & 0.900          & \textbf{0.900} & \textbf{0.871} & 0.761          & 0.519          & 0.552          & 0.627          & 0.678 \\ [3pt]
\hline
\emph{Lesion-Net (Dual Dice loss)} \\
Lesion-Net-32s      && 0.573          & 0.289          & 0.590          & 0.730          & 0.539          & 0.632          & 0.536          & 0.582   & \textbf{0.687}
                    && 0.792          & 0.899          & 0.881          & 0.857          & 0.778          & 0.720          & \textbf{0.773} & 0.669          & 0.762 \\
Lesion-Net-16s      && 0.591          & 0.377          & 0.612          & 0.740          & 0.565          & 0.645          & 0.590          & 0.571         & 0.623
                    && \textbf{0.801} & 0.902          & 0.882          & 0.866          & 0.792          & 0.733          & 0.726          & 0.701   & \textbf{0.807}\\
Lesion-Net-8s       && \textbf{0.603} & 0.377          & \textbf{0.617} & 0.740          & 0.575          & \textbf{0.648} & \textbf{0.616} & 0.580          & 0.667 
                    && 0.780          & 0.900          & 0.881          & 0.861          & 0.771          & 0.687          & 0.693          & 0.711          & 0.733 \\
Lesion-Net-4s       && 0.592          & 0.394          & 0.614          & 0.743          & \textbf{0.577} & 0.633          & 0.588          & 0.570          & 0.616
                    && 0.781          & \textbf{0.904} & 0.883          & 0.862          & 0.791          & 0.678          & 0.660          & \textbf{0.719} & 0.748 \\
Lesion-Net-2s       && 0.581          & 0.381          & 0.614          & \textbf{0.744} & 0.567          & 0.634          & 0.569          & 0.565          & 0.572
                    && 0.787          & 0.900          & 0.893          & 0.868          & \textbf{0.793} & 0.706          & 0.706          & 0.667          & 0.764 \\

\hline
Lesion-Net-16s (WCE)     && 0.364            & 0.180          & 0.389          & 0.543         & 0.346            & 0.424         & 0.245        & 0.363          & 0.423  
                    && 0.534            & 0.864         & 0.822         & 0.780         & 0.574             & 0.324         & 0.274         & 0.354         & 0.282\\
Lesion-Net-16s (Focal)   && 0.458             & 0.165         & 0.479         & 0.682         & 0.409             & 0.471         & 0.488         & 0.423         & 0.548
                    && 0.745             & 0.880         & 0.869         & 0.837         & 0.723             & 0.650         & 0.693         & 0.627         & 0.683\\
Lesion-Net-16s (Dice)    && 0.594          & 0.362          & 0.609          & 0.734          & 0.570          & 0.637          & 0.587          & 0.573          & 0.683 
                    && 0.769          & 0.899          & 0.860          & 0.851          & 0.754          & 0.709          & 0.661          & 0.694          & 0.724 \\

\bottomrule
\end{tabular}}
\vspace{-5mm}
\end{table*}

\textbf{Baselines}. Our criteria for choosing baselines are two-fold: state-of-the-art in related tasks and open-source, allowing us to run them with the same preciseness as intended by their developers. Four prior arts, \ie, FCN \cite{fcn}, U-Net~\cite{unet}, DeepLabv3+~\cite{deeplabv3p}, 
and DANet~\cite{cvpr19-danet}, are compared. For all networks, we use Inception-v3 as the backbone of their contracting paths.
%
In addition, as the majority of the existing works utilize a patch-based sliding window approach to detect retinal lesions, we include patch-based FCN-32s. To train the patch-based model, we uniformly divide each image into $4\times4$ patches, each sized to 224$\times$224. Given a test image, the model is run with a window size of $224\times224$ and a stride of $112$. Scores from overlapped areas are averaged. 
All the baselines are trained with the Dice loss. 



\textbf{Comparing Lesion-Net with distinct settings}. 
As Table \ref{tab:lesion-seg} shows, the overall performance of Lesion-Net increases first, from 0.573 (Lesion-Net-32s) to 0.591 (Lesion-Net-16s), and decreases later, from 0.592 (Lesion-Net-4s ) to 0.581 (Lesion-Net-2s). The peak is obtained by Lesion-Net-8s, with an F1 of 0.603. The result confirms our hypothesis that when the network parameters keep increasing, the additional layers can have a negative effect on the performance.


\textbf{Comparing loss functions}. 
As the parameter $\alpha$ in Focal loss~\cite{iccv17-focal} is dataset-dependent, we set it to $0.8$ according to our validation set, with the parameter $\gamma$ set to $2$ as suggested in the original paper. Dice and the proposed dual loss outperform Focal and WCE\cite{gdice} with a large margin, see Table \ref{tab:lesion-seg}. Correcting small-sized errors cannot be well reflected by the pixel-wise F1 score. This explains the relatively small difference between Dice and the dual loss for lesion segmentation.



\begin{figure*} [tbh!]
\centering
\includegraphics[width=1.8\columnwidth]{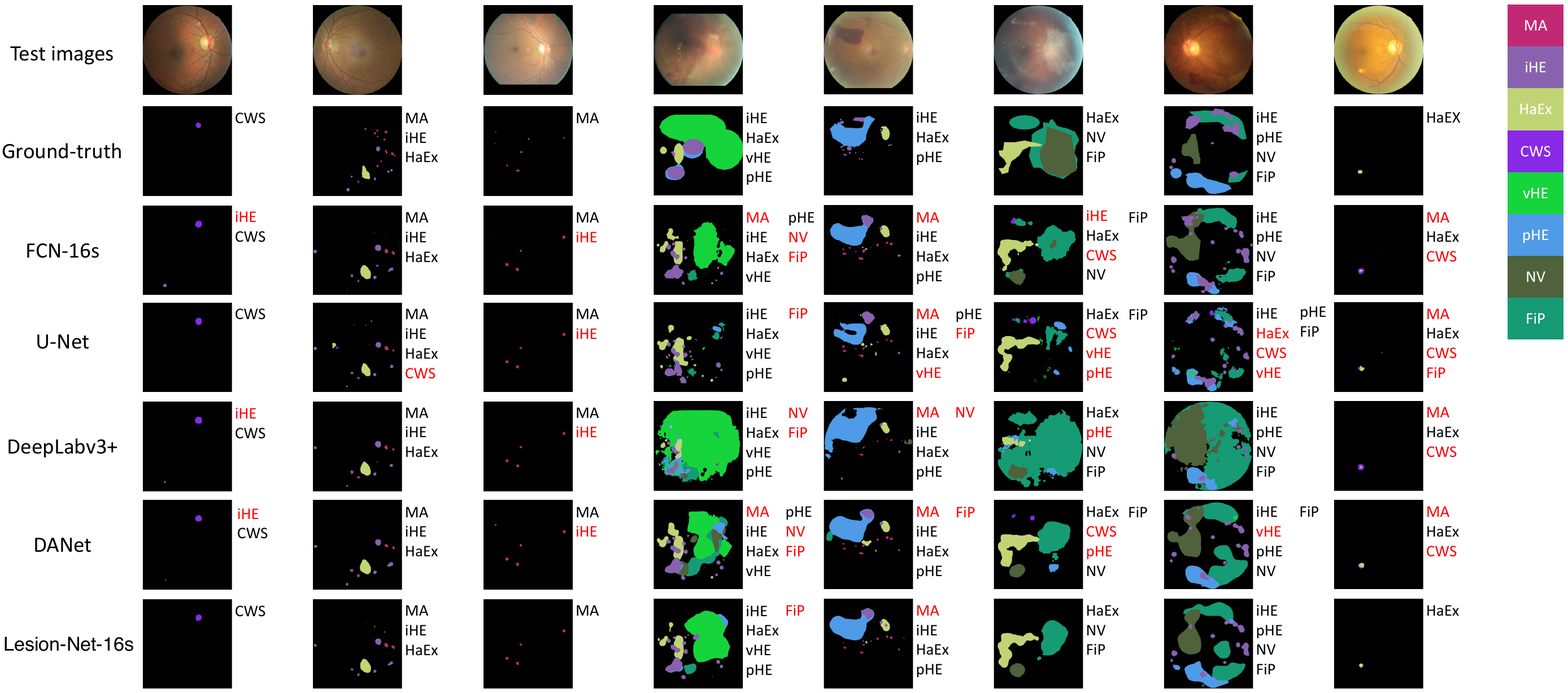}
\vspace{-3mm}
\caption{
\textbf{Some qualitative results of lesion segmentation and classification}. Red font indicates false alarms. The proposed Lesion-Net-16s produces more smooth segmentation boundaries and less false alarms. Best viewed digitally.}
\label{fig:sup_example}
\vspace{-6mm}
\end{figure*}

\textbf{Comparing with the baselines}. Lesion-Net outperforms the baselines. 
Patch-based FCN-32s is less effective than its full-resolution counterpart. As noted in Section \ref{sec:related}, properly recognizing MA requires a holistic view, which is absent for the patch-based model. This explains its lowest performance (F1 of 0.209) on this lesion. Patch-based FCN-32s also has difficulty in segmenting large lesions such as vHE and pHE.
Compared to DeepLabv3+, Lesion-Net shows similar performance on MA, iHE, HaEx and CWS while noticeably better for vHE, pHE, NV and FiP. Comparing the two groups of lesions, the latter lack clear boundaries. As shown in Fig. \ref{fig:sup_example}, irregular segmentation boundaries produced by DeepLabv3+ implies its attempt to produce precise boundaries, which are however unnecessary for retinal lesions. The results confirm our hypothesis that DeepLabv3+ is over-designed for this task. In the meantime, the viability of the proposed Lesion-Net for retinal lesion segmentation is justified.

\begin{figure} [tbh!]
\centering
\includegraphics[width=0.9\columnwidth]{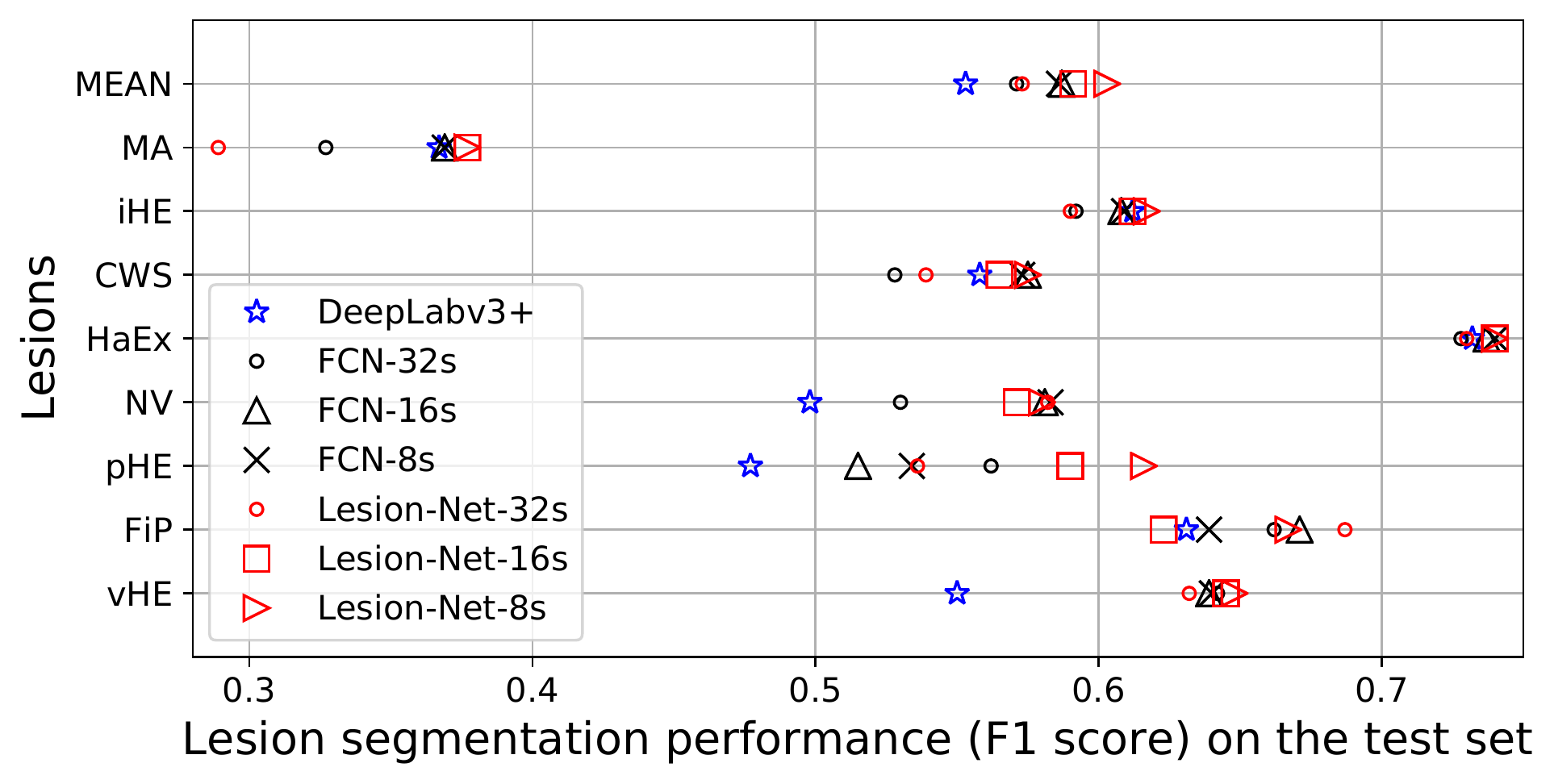}
\vspace{-3mm}
\caption{Illustrating Table \ref{tab:lesion-seg}, lesions sorted by their average size.}
\label{fig:seg-perf}
\vspace{-4mm}
\end{figure}

As shown in Fig. \ref{fig:seg-perf}, for small-sized lesions with relatively clear boundaries (MA, iHE, CWS and HaEx), we observe close performance among distinct models. Exceptions are L-Net-32s and FCN-32s, as they do 32$\times$ upsampling by parameter-free bilinear interpolation, and thus difficult to accurately locate small lesions. For large lesions yet with imprecise boundaries (NV, pHD and vHE), the simplicity of FCN and L-Net becomes advantageous. The L-Net series produce more smooth segmentation boundaries. The fact that the top performer for FiP is L-Net-32s is due to the relatively clear boundary of this large lesion.

\subsection{Experiment 2. Lesion Classification} \label{sssec:exp-lesion-det}

\textbf{Baselines}. We re-use the baselines from Experiment 1, with lesion classification obtained by global max pooling on segmentations. We also compare with two segmentation-free models, \ie, Inception-v3~\cite{inception-v3} and ABN~\cite{abn}, both trained using image-level lesion annotations and Dice.



\textbf{Comparing Lesion-Net with distinct settings}. As Table \ref{tab:lesion-seg} shows, for lesion classification Lesion-Net with a shorter expansive path, \eg~Lesion-Net-16s and Lesion-Net-32s, is preferred. From Table \ref{tab:lesion-seg} we see that Lesion-Net trained with the dual loss is the best, suggesting small misclassified blobs are reduced.

\textbf{Comparing with the baselines}. Inception-v3 and ABN are less effective than the majority of the segmentation based models. The results suggest the importance of lesions' spatial information even for making image-level predictions. Different from its behavior for lesion segmentation, DeepLabv3+ becomes runner-up for lesion classification. For vHE, this model outperforms the others with a large margin. Note that DeepLabv3+ is specifically designed to capture multi-scale information by its parallel dialated convolutions. This design appears to be good at capturing the major pattern of vHE which often occupies more than half of an image. Overall Lesion-Net-16s is the best.

\subsection{Experiment 3. Lesions for DR Grading} \label{sssec:exp-dr}

\textbf{Baselines}. We again compare with Inception-v3 and ABN, both re-trained for DR grading.
One might also consider a more straightforward method that enriches the output of the GAP layer by concatenating the $m$-dimensional lesion vector $(P_1, \ldots, P_m)$. Accordingly, the size of the fully connected layer is adjusted to $(2,048+m) \times 5$. Note that similar ideas have been exploited in the context of image captioning for obtaining semantically enhanced image features~\cite{tmm2019-cococn}. We term this baseline Lesion-Concat.


\textbf{Results}. We use Lesion-Net-16s in the multi-task network for its best overall performance in the previous experiments. As Table \ref{tab:grading} shows, using a better lesion segmentation model results in more accurate DR grading,
with the multi-task network (Lesion-Net-16s) as the top performer.
The better performance of learned weights (Eq. \ref{eq:att-w2}) compared to CW-MaxPool (Eq. \ref{eq:att-w1}) supports our statement that a region deemed to be negative with respect to the lesions does not necessarily mean it is useless for DR grading



\begin{table} [tb!]
\renewcommand{\arraystretch}{1.1}
\centering
\caption{\textbf{DR grading results}. Numbers after Inception-v3 means the input resolution.}
\label{tab:grading}
\vspace{-3mm}
\scalebox{0.78}{
\begin{tabular}{@{}ll l l r@{}}
\toprule
\textbf{DR model} & \textbf{Attention weights}  &  \textbf{Lesion model}  && \textbf{Kappa}\\
\hline
Inception-v3 ($224\times 224$)  & --                                & --                && 0.660 \\
Inception-v3 ($488\times 448$)  & --                                & --                && 0.729 \\
Inception-v3                    & --                                & --                && 0.774 \\
Lesion-Concat                   & --                                & Inception-v3      && 0.780 \\
\emph{Multi-task network}       & Conv (Eq. \ref{eq:att-w2})        & U-Net             && 0.780 \\
\emph{Multi-task network}       & CW-MaxPool (Eq. \ref{eq:att-w1})  & Lesion-Net-16s    && 0.781 \\
\emph{Multi-task network}       & Conv (Eq. \ref{eq:att-w2})        & FCN-8s            && 0.787 \\
\emph{Multi-task network}       & Conv (Eq. \ref{eq:att-w2})        & DeepLabv3+        && 0.787 \\
Lesion-Concat                   & --                                & Lesion-Net-16s    && 0.788 \\
ABN-grading                     & --                                & --                && 0.797 \\
\emph{Multi-task network}       & Conv (Eq. \ref{eq:att-w2})        & Lesion-Net-16s    && \textbf{0.803}\\
\bottomrule
\end{tabular}}
\vspace{-6mm}
\end{table}



We summarize the performance in Table \ref{tab:perf-overview}. When compared to the best baseline per task, the improvement seems to be not significant. However, for the best overall performance, one has to simultaneously deploy three distinct baselines (FCN-8s, DeepLabv3+ and ABN-grading) with 3.2GB GPU memory at run time, while our multi-task network performs better in all three tasks with half GPU memory (1.5GB).

\begin{table} [tb!]
\renewcommand{\arraystretch}{1.1}
\centering
\caption{\textbf{Overall performance of different models}.}
\label{tab:perf-overview}
\vspace{-3mm}

\scalebox{0.78}{
\begin{tabular}{@{}ll l l r@{}}
\toprule
\textbf{Model}   &  \textbf{Lesion segmentation}  & \textbf{Lesion Classification} & \textbf{DR grading} \\
\hline
FCN-8s                           & 0.586 & 0.778 & - \\
DeepLabv3+                        & 0.553 & 0.794 & - \\
ABN-grading                       & -   & -   & 0.797 \\
\emph{Our model}  & \textbf{0.591} & \textbf{0.801} & \textbf{0.803} \\
\bottomrule
\end{tabular}}
\vspace{-6mm}
\end{table}



\section{Conclusions} \label{sec:conc}

We have developed a multi-task deep learning approach to lesion segmentation, lesion classification and disease classification for color fundus images. Extensive experiments justify 
the superiority of the proposed approach against the prior art. The proposed Lesion-Net, with its re-designed expansive path and the proposed dual loss, is found to be effective for learning from retinal lesion annotations with imprecise boundaries. Exploiting Lesion-Net as a side-attention branch, the multi-task network simultaneously improves DR grading and interprets the decision with lesion maps. 


While working on fundus images, our work reveals good practices for developing a semantic segmentation network given training data with imprecise object boundaries and extremely imbalanced classes, and for converting attributes predicted at pixel-level to categories at a higher level. We believe the lessons learned are beyond the specific domain.  

\section*{Acknowledgments}
This research was supported in part by the National Natural Science Foundation of China (No. 61672523), Beijing Natural Science Foundation (No. 4202033), Beijing Natural Science Foundation Haidian Original Innovation Joint Fund (No. 19L2062), the Non-profit Central Research Institute Fund of Chinese Academy of Medical Sciences (No. 2018PT32029), CAMS Initiative for Innovative Medicine (CAMS-I2M, 2018-I2M-AI-001), and the Pharmaceutical Collaborative Innovation Research Project of Beijing Science and Technology Commission (No. Z191100007719002). Corresponding authors: Xirong Li and Youxin Chen.

\clearpage






%

\bibliographystyle{IEEEtran}
\bibliography{ref}

\end{document}